\documentclass[sn-nature]{sn-jnl}

\usepackage{graphicx}%
\usepackage{multirow}%
\usepackage{amsmath,amssymb,amsfonts}%
\usepackage{amsthm}%
\usepackage{mathrsfs}%
\usepackage[title]{appendix}%
\usepackage{xcolor}%
\usepackage{textcomp}%
\usepackage{manyfoot}%
\usepackage{booktabs}%
\usepackage{algorithm}%
\usepackage{algorithmicx}%
\usepackage{algpseudocode}%
\usepackage{listings}%
\usepackage{siunitx}
\usepackage{float}
\usepackage{makecell}

\theoremstyle{thmstyleone}%
\theoremstyle{thmstyletwo}%

\theoremstyle{thmstylethree}%
\begin{document}

\title[Article Title]{
Pseudodata-guided Invariant Representation Learning Boosts the Out-of-Distribution Generalization in Enzymatic Kinetic Parameter Prediction 
}

\author[1,2]{\fnm{Haomin} \sur{Wu}}
\equalcont{These authors contributed equally to this work.}
\author*[3]{\fnm{Zhiwei} \sur{Nie}}\email{zhiweiNie@pku.edu.cn}
\equalcont{These authors contributed equally to this work.}

\author[3]{\fnm{Hongyu} \sur{Zhang}}
\author*[2]{\fnm{Zhixiang} \sur{Ren}}\email{jason.zhixiang.ren@outlook.com}

\affil[1]{Pengcheng Laboratory, Shenzhen, China}
\affil[2]{Southern University of Science and Technology, Shenzhen, China}
\affil[3]{School of Electronic and Computer Engineering, Peking University, Shenzhen, China}

\abstract{
Accurate prediction of enzyme kinetic parameters is essential for understanding catalytic mechanisms and guiding enzyme engineering. 
However, existing deep learning–based enzyme–substrate interaction (ESI) predictors often exhibit performance degradation on sequence-divergent, out-of-distribution (OOD) cases, limiting robustness under biologically relevant perturbations. 
We propose $\text{O}^2$DENet, a lightweight, plug-and-play module that enhances OOD generalization via biologically and chemically informed perturbation augmentation and invariant representation learning. 
$\text{O}^2$DENet introduces enzyme-substrate perturbations and enforces consistency between original and augmented enzyme-substrate-pair representations to encourage invariance to distributional shifts. 
When integrated with representative ESI models, $\text{O}^2$DENet consistently improves predictive performance for both $k_{cat}$ and $K_m$ across stringent sequence-identity–based OOD benchmarks, achieving state-of-the-art results among the evaluated methods in terms of accuracy and robustness metrics.
Overall, $\text{O}^2$DENet provides a general and effective strategy to enhance the stability and deployability of data-driven enzyme kinetics predictors for real-world enzyme engineering applications.
}

\maketitle

\section{Introduction}\label{sec1}

Accurately modeling enzyme-substrate interactions (ESIs) is fundamental for understanding biochemical reaction mechanisms and accelerating enzyme engineering.
Recent advances in deep learning have enabled substantial progress in data-driven ESI prediction, with models such as UniKP~\cite{UniKP}, DLKCat~\cite{DLKCat}, CatPred~\cite{CatPred}, and OmniESI~\cite{OmniESI} leveraging convolutional networks, graph neural networks~\cite{gnn,gnn1}, or pretrained protein language models~\cite{plm1,plm2,plm3,plm4} and molecular models~\cite{mm1, mm2, mm3}.
For example, enzyme kinetic parameters enzyme turnover number ($k_{cat}$)~\cite{kcat} and Michaelis constant ($K_{m}$)~\cite{Km}, as important indicators of enzyme-catalyzed specific reactions~\cite{ecsr}, can be predicted by deep learning models, rather than by time-consuming, expensive, and labor-intensive experimental measurements.

Despite these advances, a persistent challenge remains: the severe distribution shift between training and deployment environments, particularly when predicting interactions for enzymes with low sequence similarity to those seen during training, i.e., out-of-distribution (OOD) generalization scenarios in practical enzyme engineering~\cite{OOD1, OOD2, OOD3}.
That is, enzyme families often exhibit high sequence diversity, and newly discovered or engineered enzymes frequently fall far outside the training distribution of existing datasets.
Some evaluations based on sequence-identity clustering, such as those implemented in the CatPred-kcat and CatPred-km benchmarks~\cite{CatPred}, have revealed that current ESI models suffer notable performance degradation as sequence similarity decreases~\cite{sic1, sic2}.
This performance drop highlights a crucial bottleneck: modern ESI predictors tend to overfit to training distributions and fail to capture the invariance to biologically meaningful perturbations, limiting their robustness and real-world utility.
Therefore, a unified strategy to improve the generalization ability of ESI predictors is urgently needed.

In this work, we introduce $\text{O}^2$DENet, a lightweight and plug-and-play module designed to enhance the OOD generalization of ESI predictors through pseudodata-augmented invariant learning~\cite{ivl, pd}.
Rather than modifying network backbones, $\text{O}^2$DENet operates by augmenting both enzyme and substrate representations to expand training distribution and enforcing feature-level consistency to capture task-relevant invariant feature.
First, enzyme sequences undergo randomly residue masking, while substrate molecules are perturbed using SMILES enumeration and chemically constrained graph masking.
The above multi-perspective pseudodata augmentation generates diverse enzyme-substrate pairs that may have similar molecular interactions, which expands the training distribution available to the deep-learning model.
Then, invariant learning on these augmented enzyme-substrate pairs prompts the predictive model to learn perturbation-invariant latent representations, thereby improving its ESI prediction performance in OOD scenarios.

Integrated seamlessly into multiple representative ESI predictive frameworks, $\text{O}^2$DENet consistently and significantly improved their generalization performance across different OOD levels, ranging from 40\% to 99\% sequence identity, without additional architectural complexity.
These improvements demonstrate that carefully designed pseudodata augmentations, combined with invariance learning, offered an efficient and effective solution to the long-standing OOD generalization challenge in enzyme--substrate prediction.
Overall, by enabling ESI predictive models to generalize more effectively to unseen enzyme families, $\text{O}^2$DENet represents a step toward more reliable biochemical prediction systems suitable for real-world deployment.

\section{Results}

\subsection{The architecture of $\text{O}^2$DENet}

\begin{figure}[h!]
  \centering
  \includegraphics[width=1.05\textwidth]{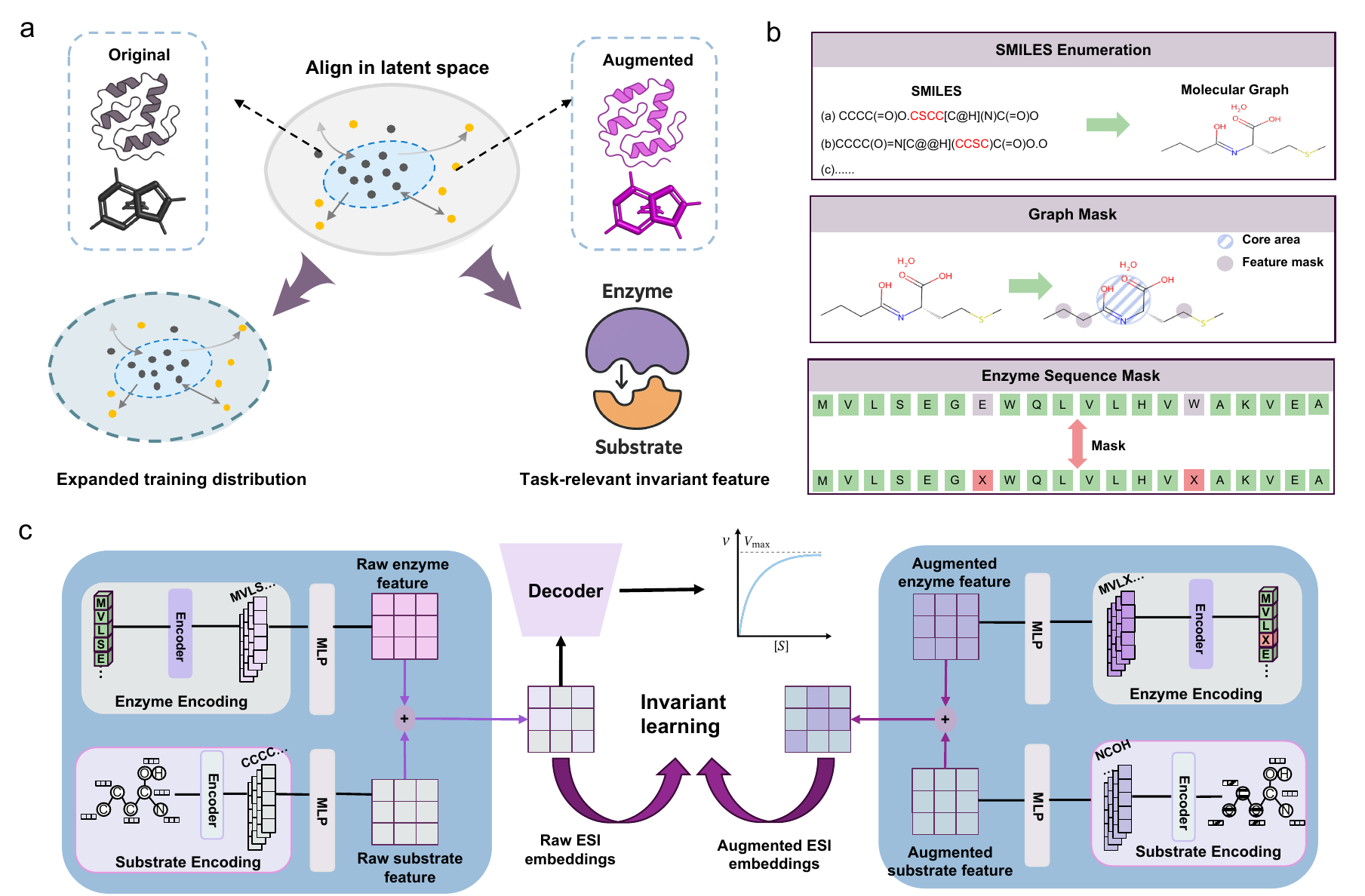} 
  \caption{The motivation and methodology of $\text{O}^2$DENet.
  \textbf{a}, The methodology of $\text{O}^2$DENet for improving the generalization ability of enzyme-substrate interaction predictors.
  \textbf{b}, The augmentation approaches of enzyme and substrate inputs, namely SMILES enumeration, molecular graph masking, and enzyme sequence mask.
  \textbf{c}, The implementation of $\text{O}^2$DENet, where augmented ESI embedding undergoes invariant learning with the raw ESI embedding as an auxiliary training task.
  }
  \label{fig1}
\end{figure}

In practical enzyme engineering, ESI predictors face a severe out-of-distribution generalization challenge, that is, a significant decrease in predictive performance due to the studied enzyme families dissimilar to the training data.
As shown in Fig.\ref{fig1}a, this fragility in model generalization stems from the insufficient distribution of the training data and the failure of neural networks to internalize invariant features across different enzyme-substrate pairs relevant to the ESI prediction task.
Correspondingly, $\text{O}^2$DENet addresses the above challenges through pseudodata-augmented invariant learning.

First, $\text{O}^2$DENet improves the generalization ability of ESI predictors by augmenting enzyme and substrate inputs to construct pseudo-enzyme-substrate pairs.
Specifically, as illustrated in Fig.\ref{fig1}b, substrate molecules are augmented either by SMILES enumeration or by chemically constrained masking on the 2D molecular graph.
SMILES enumeration is designed to accommodate encoders trained with SMILES, whereas graph masking is applied to align with graph-based encoders.
Notably, in the graph-masking strategy, key functional groups are preserved as the core region, while the remaining atoms are randomly masked at a predefined ratio. 
For enzyme sequences, residues are similarly masked at a certain ratio. 
These augmented enzyme–substrate pairs are expected to retain catalytic interactions analogous to those of the raw pairs, thereby expanding the neighborhood of the training distribution and ultimately reducing the model’s generalization error.

Second, $\text{O}^2$DENet improves the generalization ability of ESI predictors by performing invariant learning between pseudo-enzyme-substrate pairs and the raw pairs.
Specifically, as illustrated in Fig.\ref{fig1}c, the respective embeddings of augmented pseudo-enzyme-substrate pair are concatenated into an augmented ESI embedding. 
This augmented ESI embedding, along with the raw ESI embedding, undergoes invariant learning to constrain consistency at the feature level.
It is worth noting that the above invariant learning is an auxiliary training task; only the ESI embeddings corresponding to the raw enzyme-substrate pairs are encoded for the actual downstream tasks, namely $k_{cat}$ and $K_{m}$ prediction.
This invariant learning between the pseudo-enzyme-substrate pair and the raw pair forces the ESI predictors to capture perturbation-invariant features related to the catalytic reaction and filter out task-irrelevant noise, thereby improving the model's generalization performance.

In summary, the parameter-lightweight $\text{O}^2$DENet requires no modification to the network architecture of existing ESI predictors, which only augments the input enzyme-substrate pairs and adds an invariant-learning auxiliary training task, thus enabling plug-and-play application to various ESI predictive frameworks.

\subsection{Performance evaluation of $\text{O}^2$DENet on top of various ESI predictors}

To evaluate the universality and effectiveness of $\text{O}^2$DENet, we integrated it with four representative ESI predictive frameworks, namely DLKCat, UniKP, CatPred, and OmniESI.
To fully evaluate the improvement in generalization performance of $\text{O}^2$DENet, we adopted the CatPred database, which contains four different enzyme-sequence-identity test sets (40\%, 60\%, 80\% and 99\%) representing different OOD levels.
In this work, the predictions of two enzyme kinetic parameters, namely $k_{cat}$ and $K_{m}$, were set as downstream tasks to evaluate the effectiveness of $\text{O}^2$DENet.
For all baselines, our $\text{O}^2$DENet was incorporated without modifying the backbone network or task-specific loss design; only an additional feature consistency regularization term was applied during training.

\begin{table}[h!]
\centering
\begin{tabular}{c|cccc}
\hline
\multirow{2}{*}{\textbf{Model}} & \multicolumn{4}{c}{$\pmb{R^2~\uparrow}$} \\
& OOD--99\% & OOD--80\% & OOD--60\% & OOD--40\% \\
\hline
DLKCat 
& 0.117 & 0.108 & 0.085 & 0.102 \\
DLKCat{+}$\text{O}^2$DENet 
& \textbf{0.210 (+79.5\%)} & \textbf{0.265 (+145.4\%)} & \textbf{0.136 (+60.0\%)} & \textbf{0.132 (+29.4\%)} \\
UniKP 
& 0.372 & 0.349 & 0.302 & 0.260 \\
UniKP{+}$\text{O}^2$DENet 
& \textbf{0.408 (+9.7\%)} & \textbf{0.378 (+8.3\%)} & \textbf{0.362 (+19.9\%)} & \textbf{0.312 (+20.0\%) }\\
CatPred 
& 0.404 & 0.333 & 0.355 & 0.365 \\
CatPred{+}$\text{O}^2$DENet 
& \textbf{0.406 (+0.5\%)} & \textbf{0.391 (+17.4\%)} & \textbf{0.377 (+6.2\%)} & \textbf{0.368 (+0.8\%)} \\
OmniESI 
& 0.409 & 0.401 & 0.356 & 0.342 \\
OmniESI{+}$\text{O}^2$DENet 
& \textbf{0.475 (+16.1\%)} & \textbf{0.465 (+16.0\%)} & \textbf{0.478 (+34.3\%)} & \textbf{0.410 (+19.9\%)} \\
\hline
\end{tabular}
\caption{$R^2$ values on different OOD test sets for the $k_{cat}$ prediction task. 
Results after integrating $\text{O}^2$DENet are highlighted in bold, with relative improvements over the baselines in parentheses.}
\label{table:kcat-r2}
\end{table}

\begin{table}[h!]
\centering
\begin{tabular}{c|cccc}
\hline
\multirow{2}{*}{\textbf{Model}} & \multicolumn{4}{c}{\textbf{MAE $\pmb{\downarrow}$}} \\
& OOD--99\% & OOD--80\% & OOD--60\% & OOD--40\% \\
\hline
DLKCat 
& 1.523 & 1.518 & 1.583 & 1.556 \\
DLKCat{+}$\text{O}^2$DENet 
& \textbf{1.470 (-3.5\%)} 
& \textbf{1.438 (-5.3\%)} 
& \textbf{1.512 (-4.5\%)} 
& \textbf{1.503 (-3.4\%)} \\
UniKP 
& 1.043 & 1.071 & 1.179 & 1.189 \\
UniKP{+}$\text{O}^2$DENet 
& \textbf{0.992 (-4.9\%)} 
& \textbf{1.065 (-0.6\%)} 
& \textbf{1.120 (-5.0\%)} 
& \textbf{1.012 (-14.9\%)} \\
CatPred 
& 0.987 & 1.029 & 1.124 & 1.131 \\
CatPred{+}$\text{O}^2$DENet 
& \textbf{0.963 (-2.4\%)} 
& \textbf{1.007 (-2.1\%)} 
& \textbf{1.069 (-4.9\%)} 
& \textbf{1.103 (-2.5\%)} \\
OmniESI 
& 0.956 & 0.989 & 1.062 & 1.142 \\
OmniESI{+}$\text{O}^2$DENet 
& \textbf{0.920 (-3.8\%)} 
& \textbf{0.929 (-6.1\%)} 
& \textbf{0.935 (-12.0\%)} 
& \textbf{0.931 (-18.5\%)} \\
\hline
\end{tabular}
\caption{MAE values on different OOD test sets for the $k_{cat}$ prediction task. 
Results after integrating $\text{O}^2$DENet are highlighted in bold, with relative improvements over the baselines in parentheses.}
\label{table:kcat-mae}
\end{table}

\begin{table}[h!]
\centering
\begin{tabular}{c|cccc}
\hline
\multirow{2}{*}{\textbf{Model}} & \multicolumn{4}{c}{$\pmb{R^2~\uparrow}$} \\
& OOD--99\% & OOD--80\% & OOD--60\% & OOD--40\% \\
\hline
DLKCat 
& 0.383 & 0.392 & 0.397 & 0.353 \\
DLKCat{+}$\text{O}^2$DENet 
& \textbf{0.402 (+5.0\%)} & \textbf{0.413 (+5.4\%)} & \textbf{0.407 (+2.5\%)} & \textbf{0.392 (+11.0\%)} \\
UniKP 
& 0.493 & 0.496 & 0.409 & 0.449 \\
UniKP{+}$\text{O}^2$DENet 
& \textbf{0.516 (+4.7\%)} & \textbf{0.516 (+4.0\%)} & \textbf{0.413 (+1.0\%)} & \textbf{0.462 (+2.9\%) }\\
CatPred 
& 0.533 & 0.540 & 0.539 & 0.475 \\
CatPred{+}$\text{O}^2$DENet 
& \textbf{0.537 (+0.8\%)} & \textbf{0.544 (+0.7\%)} & \textbf{0.545 (+1.1\%)} & \textbf{0.478 (+0.6\%)} \\
OmniESI 
& 0.541 & 0.549 & 0.534 & 0.472 \\
OmniESI{+}$\text{O}^2$DENet 
& \textbf{0.576 (+6.5\%)} & \textbf{0.584 (+6.4\%)} & \textbf{0.551 (+3.2\%)} & \textbf{0.519 (+10.0\%)} \\
\hline
\end{tabular}
\caption{$R^2$ values on different OOD test sets for the $K_{m}$ prediction task. 
Results after integrating $\text{O}^2$DENet are highlighted in bold, with relative improvements over the baselines in parentheses.}
\label{table:km-r2}
\end{table}

\begin{table}[h!]
\centering
\begin{tabular}{c|cccc}
\hline
\multirow{2}{*}{\textbf{Model}} & \multicolumn{4}{c}{\textbf{MAE $\pmb{\downarrow}$}} \\
& OOD--99\% & OOD--80\% & OOD--60\% & OOD--40\% \\
\hline
DLKCat 
& 1.261 & 1.259 & 1.280 & 1.291 \\
DLKCat{+}$\text{O}^2$DENet 
& \textbf{1.157 (-8.2\%)} 
& \textbf{1.167 (-7.3\%)} 
& \textbf{1.070 (-16.4\%)} 
& \textbf{1.130 (-12.5\%)} \\
UniKP 
& 0.688 & 0.708 & 0.731 & 0.800 \\
UniKP{+}$\text{O}^2$DENet 
& \textbf{0.658 (-4.4\%)} 
& \textbf{0.662 (-6.5\%)} 
& \textbf{0.717 (-1.9\%)} 
& \textbf{0.732 (-8.5\%)} \\
CatPred 
& 0.653 & 0.665 & 0.693 & 0.771 \\
CatPred{+}$\text{O}^2$DENet 
& \textbf{0.647 (-0.9\%)} 
& \textbf{0.661 (-0.6\%)} 
& \textbf{0.687 (-0.9\%)} 
& \textbf{0.760 (-1.4\%)} \\
OmniESI 
& 0.639 & 0.655 & 0.694 & 0.765 \\
OmniESI{+}$\text{O}^2$DENet 
& \textbf{0.606 (-5.2\%)} 
& \textbf{0.618 (-5.6\%)} 
& \textbf{0.646 (-6.9\%)} 
& \textbf{0.692 (-9.5\%)} \\
\hline
\end{tabular}
\caption{MAE values on different OOD test sets for the $K_{m}$ prediction task. 
Results after integrating $\text{O}^2$DENet are highlighted in bold, with relative improvements over the baselines in parentheses.}
\label{table:km-mae}
\end{table}

As shown in Table \ref{table:kcat-r2} and  Table \ref{table:kcat-mae}, 
for $k_{cat}$ prediction task, $\text{O}^2$DENet significantly improved the $R^2$ values of the baseline models across four different sequence-identity-based OOD test sets in most cases, with the highest relative improvement reaching 145.4\%.
At the same time, the MAE value also consistently decreased across the board with the incorporation of $\text{O}^2$DENet, with the highest relative improvement reaching 18.5\%.
For $K_{m}$ prediction task in Table \ref{table:km-r2} and Table \ref{table:km-mae}, both $R^2$ and MAE were improved across the board, with the maximum relative improvement in $R^2$ reaching 11\% and that in MAE reaching 16.4\%.
Overall, incorporating $\text{O}^2$DENet consistently led to notable overall performance improvements for all baseline models under different OOD levels.
Notably, $\text{O}^2$DENet combined with the baseline model OmniESI achieved the best generalization performance, especially achieving the highest $R^2$ and lowest MAE values on the 40\%-sequence-identity test sets, which presents the greatest OOD challenge.

AU-GOOD (Area Under the Generalisation Out-Of-Distribution curve) is a metric designed to quantify a model’s robustness under distribution shifts~\cite{AUGOOD}. 
Specifically, it evaluates model performance across multiple levels of distributional variation and computes the area under the resulting performance–shift curve, thereby capturing how well the model maintains prediction performance as it departs from the training distribution.
In this work, for enzyme–substrate interaction prediction, significant changes, such as different substrate types and enzyme families, are quite common.
AU-GOOD can effectively characterize how rapidly ESI predictor's performance degrades under these shifts, providing a more faithful measure of generalization than a single held-out test set.
Therefore, AU-GOOD was adopted by us as an additional metric for evaluating the effectiveness of $\text{O}^2$DENet.

\begin{table}[h!]
\centering
\begin{tabular}{ccc|cc}
\hline
\multirow{2}{*}{\textbf{Model}} 
& \multicolumn{2}{c|}{\textbf{$k_{cat}$ prediction}} 
& \multicolumn{2}{c}{\textbf{$K_{m}$ prediction}} \\
& $R^2$-based $\uparrow$ & MAE-based $\downarrow$
& $R^2$-based $\uparrow$ & MAE-based $\downarrow$ \\
\hline
DLKCat              & 0.202 & 1.484 & 0.433 & 1.234 \\
DLKCat{+}$\text{O}^2$DENet   
& \textbf{0.322(+59.4\%)} 
& \textbf{1.404(-5.4\%)} 
& \textbf{0.459(+6.0\%)} 
& \textbf{1.183(-4.1\%)} \\
UniKP               & 0.522 & 0.833 & 0.578 & 0.620 \\
UniKP{+}$\text{O}^2$DENet   
& \textbf{0.556(+6.5\%)} 
& \textbf{0.827(-0.7\%)} 
& \textbf{0.587(+1.6\%)} 
& \textbf{0.589(-5.0\%)} \\
CatPred             & 0.541 & 0.802 & 0.628 & 0.573 \\
CatPred{+}$\text{O}^2$DENet  
& \textbf{0.550(+1.7\%)} 
& \textbf{0.789(-1.6\%)} 
& \textbf{0.633(+0.8\%)} 
& \textbf{0.565(-1.4\%)} \\
OmniESI             & 0.569 & 0.760 & 0.639 & 0.557 \\
OmniESI{+}$\text{O}^2$DENet  
& \textbf{0.608(+6.9\%)} 
& \textbf{0.748(-1.6\%)} 
& \textbf{0.651(+1.9\%)} 
& \textbf{0.554(-0.5\%)} \\
\hline
\end{tabular}
\caption{AU-GOOD values for $k_{cat}$ and $K_{m}$ prediction tasks.
$R^2$ and MAE were used to calculate the $R^2$-based and MAE-based AU-GOOD values, respectively.
}
\label{table:au-good}
\end{table}

\begin{figure}[h!]
    \centering
    \includegraphics[width=\textwidth]{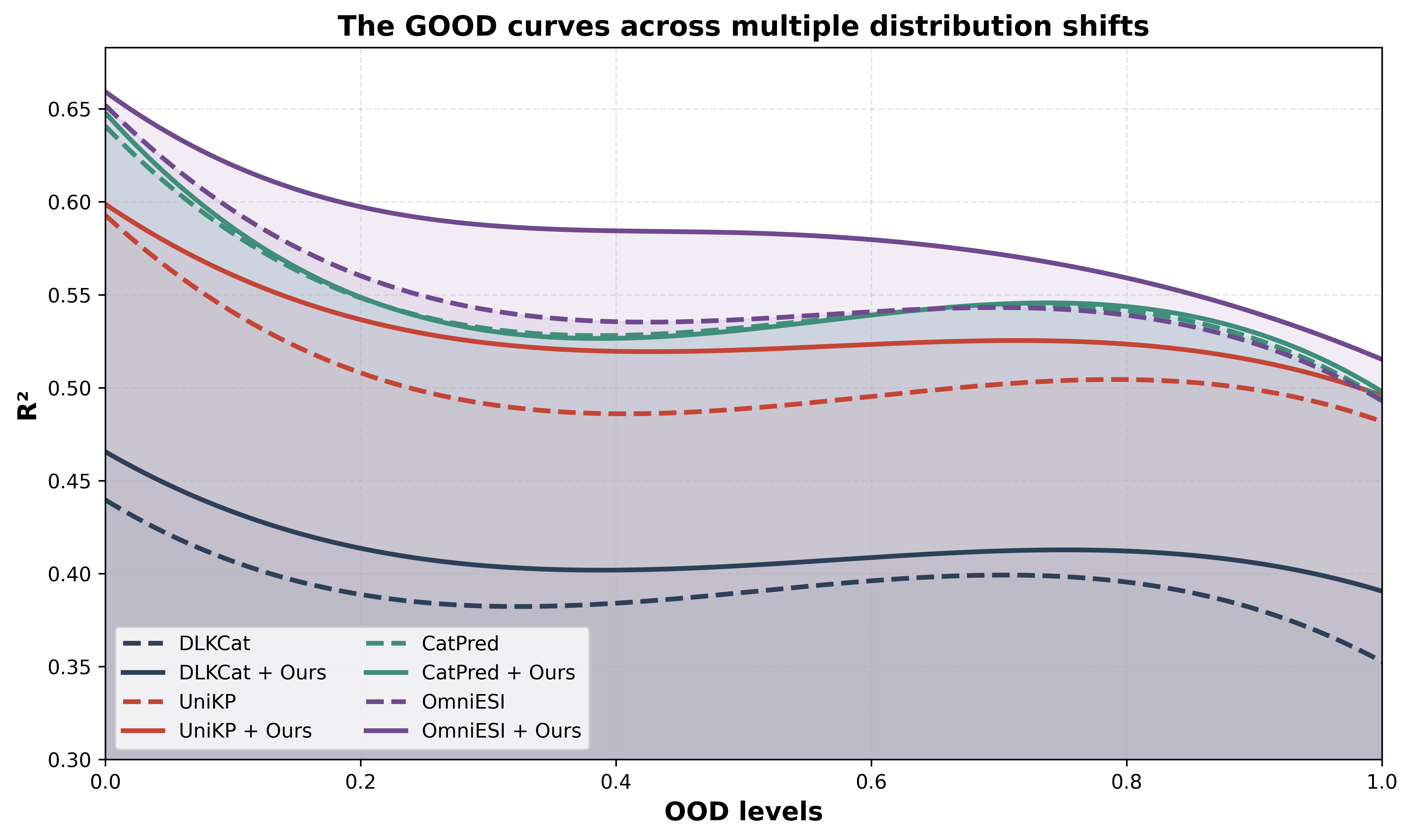}
    \caption{
    The $R^2$-based GOOD curves of the baseline models (solid lines) and their integration with $\text{O}^2$DENet (dashed lines).
    Each curve represents the trend of generalization performance under different OOD levels.
    }
    \label{fig:au-good}
\end{figure}

As shown in Table \ref{table:au-good}, the OOD generalization performance of all baseline models equipped with $\text{O}^2$DENet is comprehensively improved.
When using $R^2$ (higher is better) as the performance metric, the AU-GOOD values are improved by up to 59.4\% on the $k_{cat}$ prediction task and by 6.0\% on the $K_{m}$ prediction task.
When using MAE (lower is better) as the performance metric, the AU-GOOD values are improved by up to 5.4\% on the $k_{cat}$ prediction task and by 5.0\% on the $K_{m}$ prediction task.
To further observe the trend in the prediction performance of the baseline models due to the integration of $\text{O}^2$DENet, we plotted the $R^2$-based GOOD curves of different models in Fig.\ref{fig:au-good}, where each curve represents the trend of generalization performance under different OOD levels.
We found that models equipped with $\text{O}^2$DENet maintained higher $R^2$ values (i.e., higher $R^2$-based AU-GOOD values), and exhibit slower performance decay.
Overall, the integration of $\text{O}^2$DENet resulted in a comprehensive improvement in the generalization performance of different ESI predictors, making them more stable and robust across enzyme sequence distributional shifts.

\subsection{Ablation study for masking ratios}

For the implementation of $\text{O}^2$DENet, the substrate molecular graph and enzyme sequence are masked using predefined ratios.
To investigate the optimal masking ratio, we conducted extensive ablation experiments on top of all baseline models for $k_{cat}$ and $K_{m}$ prediction tasks.
To avoid disrupting critical structure or physicochemical properties of the enzymes and substrate molecules, we set an upper limit of the masking ratio to 30\% and an interval of 5\%.
That is, the masking ratios for the substrate molecule graph and the enzyme sequence were 5\%, 10\%, 15\%, 20\%, 25\%, and 30\%~\cite{graphmask}, respectively.

\begin{figure}[h]
  \centering
  \includegraphics[width=1.05\textwidth]{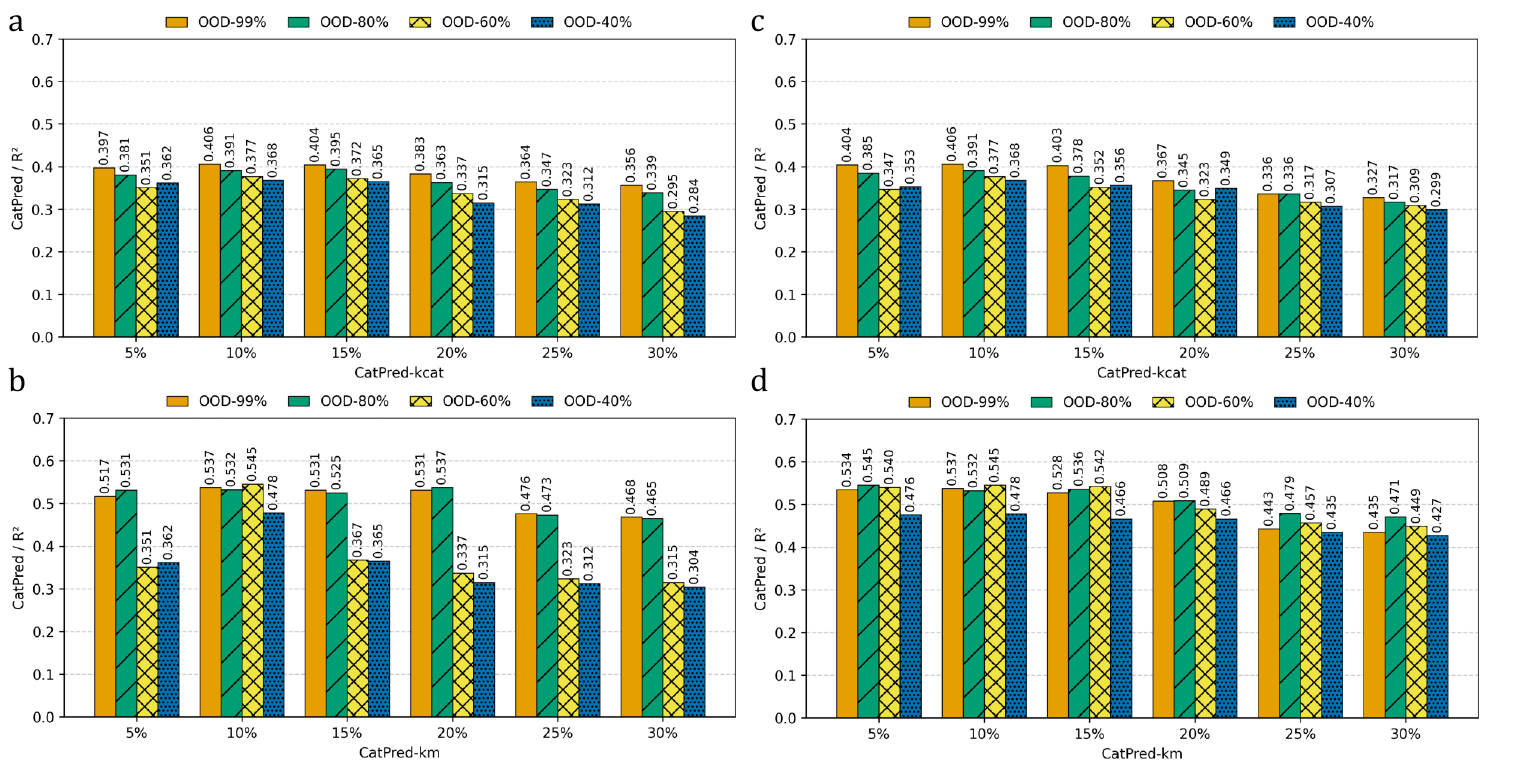} 
  \caption{
  Ablation experiments for masking ratios on top of baseline model CatPred, where $R^2$ is adopted as the evaluation metric.
  \textbf{ab}, The ablation experiments of substrate molecule graph masking ratios for $k_{cat}$ (a) and $K_{m}$ (b) prediction tasks.
  \textbf{cd}, The ablation experiments of enzyme sequence masking ratios for $k_{cat}$ (c) and $K_{m}$ (d) prediction tasks. 
  }
  \label{fig:ablation}
\end{figure}

\begin{figure}[h!]
\renewcommand{\thefigure}{S1}
  \centering
  \includegraphics[width=1.0\textwidth]{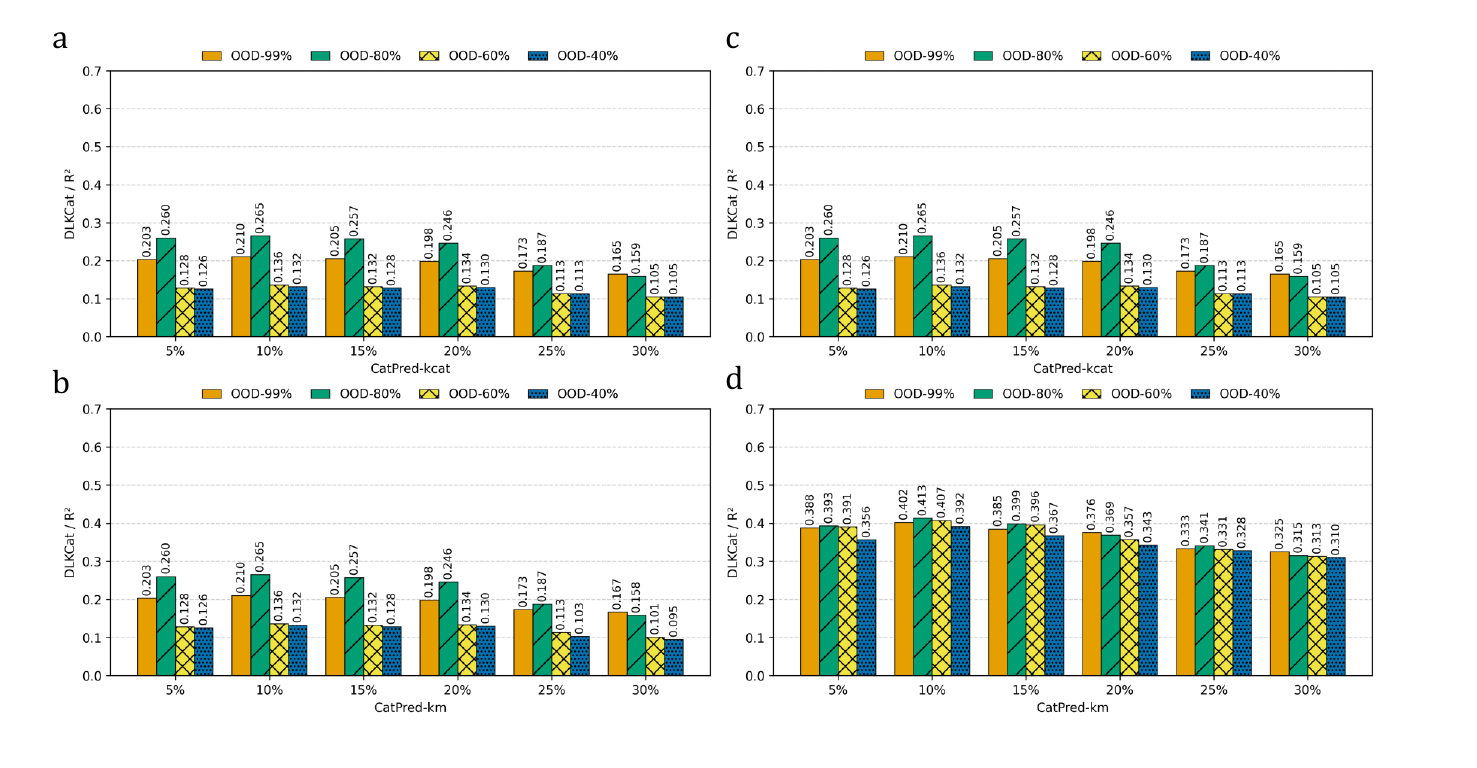} 
  \caption{
  Ablation experiments for masking ratios on top of baseline model DLKCat, where $R^2$ is adopted as the evaluation metric.
  \textbf{ab}, The ablation experiments of substrate molecule graph masking ratios for $k_{cat}$ (a) and $K_{m}$ (b) prediction tasks.
  \textbf{cd}, The ablation experiments of enzyme sequence masking ratios for $k_{cat}$ (c) and $K_{m}$ (d) prediction tasks. 
  }
  \label{figs:DLKCat_masking}
\end{figure}

\begin{figure}[h!]
\renewcommand{\thefigure}{S2}
  \centering
  \includegraphics[width=0.6\textwidth]{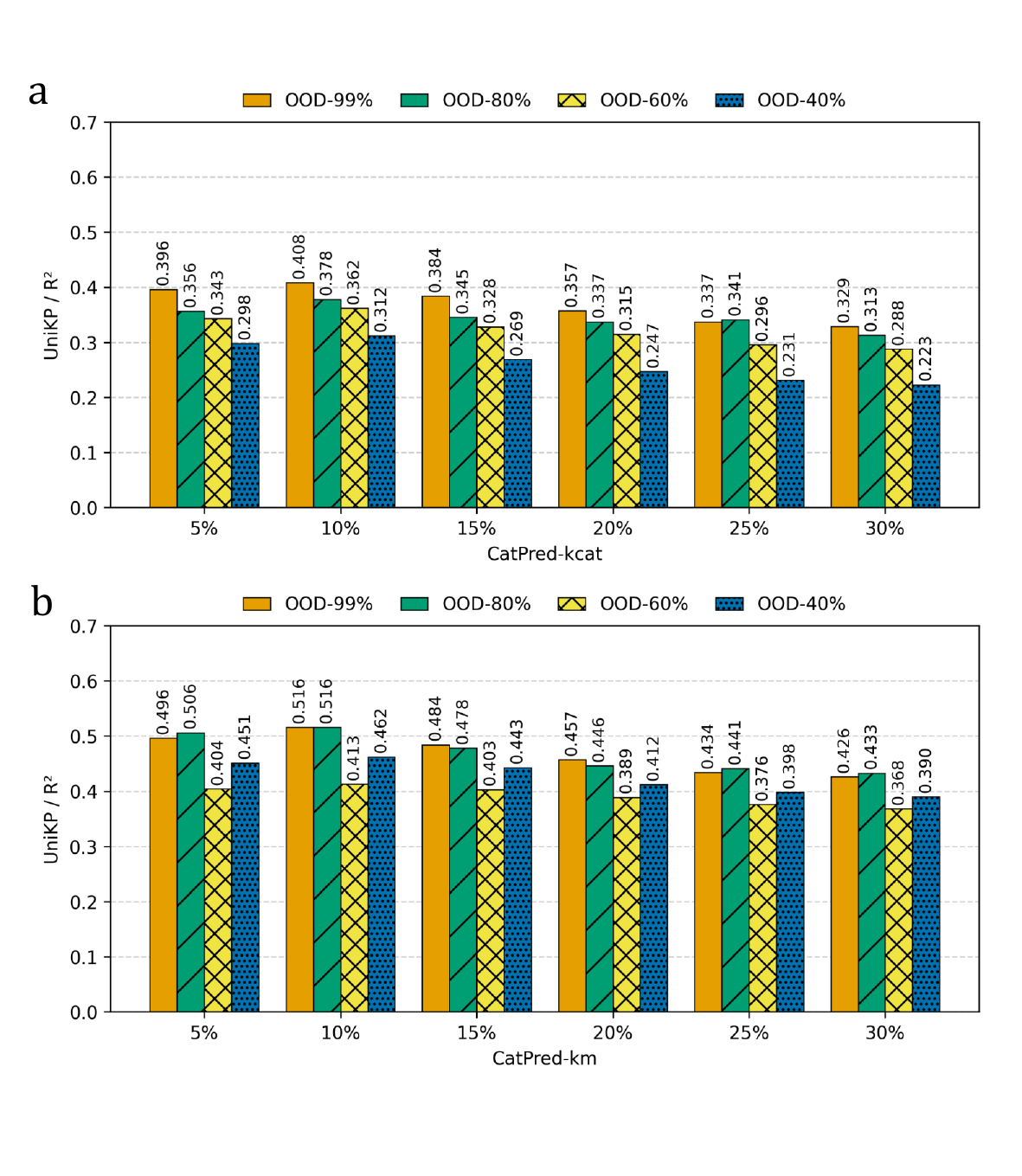} 
  \caption{
 Ablation experiments of enzyme sequence masking ratios on top of the baseline model UniKP, where $R^2$ is adopted as the evaluation metric.
\textbf{a}, The ablation experiments of enzyme sequence masking ratios for $k_{cat}$.
\textbf{b}, The ablation experiments of enzyme sequence masking ratios for $K_{m}$. 
  }
  \label{figs:UniKP_masking}
\end{figure}

\begin{figure}[h!]
\renewcommand{\thefigure}{S3}
  \centering
  \includegraphics[width=1.0\textwidth]{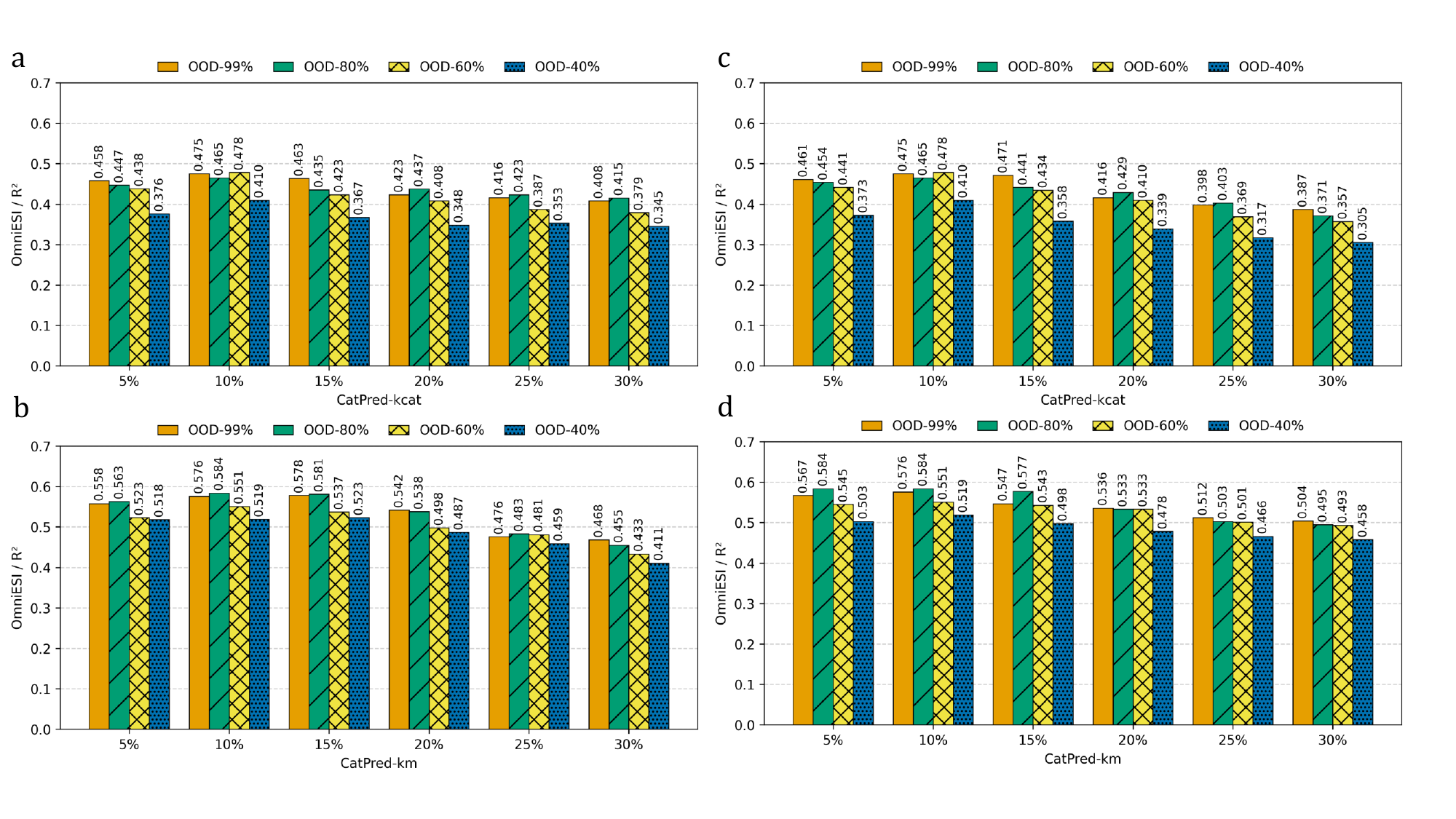} 
  \caption{
  Ablation experiments for masking ratios on top of baseline model OmniESI, where $R^2$ is adopted as the evaluation metric.
  \textbf{ab}, The ablation experiments of substrate molecule graph masking ratios for $k_{cat}$ (a) and $K_{m}$ (b) prediction tasks.
  \textbf{cd}, The ablation experiments of enzyme sequence masking ratios for $k_{cat}$ (c) and $K_{m}$ (d) prediction tasks. 
  }
  \label{figs:OmniESI_masking}
\end{figure}

As shown in Fig.\ref{fig:ablation}, ablation experiments were conducted using CatPred as the baseline model for $k_{cat}$ and $K_{m}$ prediction tasks.
For substrate molecule graph masking, a consistent increase in prediction performance was observed from 5\% to 10\% masking ratio, while the best $R^2$ values were achieved at a 10\% masking ratio.
When the masking rate increased to 15\%, the performance began to gradually deteriorated, possibly due to excessive removal of structural information or destruction of key biological properties.
Similarly, for enzyme sequence masking, the mask ratio of 10\% was also the optimal choice, resulting in the best prediction performance at most OOD levels.
The ablation results using DLKCat, UniKP, and OmniESI as baselines (Supplementary information \ref{figs:DLKCat_masking}, \ref{figs:UniKP_masking}, and \ref{figs:OmniESI_masking}) were consistent; therefore, we set the ratio of both graph mask and sequence mask to 10\% in this work.

\subsection{Zero-shot performance for practical enzyme engineering}

Modifying enzymes through directed evolution is a crucial task in enzyme engineering.
Numerous studies have explored and evolved various enzymes, but the success rate of finding candidate enzymes with superior catalytic efficiency remains low.
To validate the value of $\text{O}^2$DENet for practical enzyme engineering, we adopted two directed evolution datasets, tyrosine ammonia lyase (TAL) dataset and myrcene synthases (MS) dataset, from previously published literature~\cite{mpek}, where each variant was measured with values of $k_{cat}$ and $K_{m}$.
Subsequently, the OmniESI~\cite{OmniESI} model with the strongest predictive power, equipped with $\text{O}^2$DENet, performed zero-shot inference on each entity in both datasets to observe the prediction success rate for variants with superior catalytic efficiency.

\begin{table}[h!]
\centering
\begin{tabular}{c|ccccc}
\hline
\multirow{2}{*}{\textbf{Enzyme}} & \multicolumn{5}{c}{\textbf{TAL dataset}} \\
& Organism & Substrate & pH & Temperature & Prediction result \\
\hline
Wildtype
& Rhodotorula glutinis & L-tyrosine & 9.5 & 40 & - \\
MT-603P
& Rhodotorula glutinis & L-tyrosine & 9.5 & 40 & Correct \\
MT-366H
& Rhodotorula glutinis & L-tyrosine & 9.5 & 40 & Correct \\
MT-366W
& Rhodotorula glutinis & L-tyrosine & 9.5 & 40 & Incorrect \\
MT-587V
& Rhodotorula glutinis & L-tyrosine & 9.5 & 40 & Correct \\
MT-10Y
& Rhodotorula glutinis & L-tyrosine & 9.5 & 40 & Correct \\
MT-337C
& Rhodotorula glutinis & L-tyrosine & 9.5 & 40 & Correct \\
MT-668S
& Rhodotorula glutinis & L-tyrosine & 9.5 & 40 & Correct \\
MT-489T
& Rhodotorula glutinis & L-tyrosine & 9.5 & 40 & Incorrect \\
MT-337D
& Rhodotorula glutinis & L-tyrosine & 9.5 & 40 & Correct \\
\hline
\end{tabular}
\caption{
The zero-shot inference results of TAL dataset, where Wildtype is used as a reference.
For each variant, if our model's predicted ranking of its catalytic efficiency relative to Wildtype matches the actual ranking, it is considered a correct prediction; otherwise, it is considered incorrect.
}
\label{table:case-1}
\end{table}

\begin{table}[h!]
\centering
\begin{tabular}{c|ccccc}
\hline
\multirow{2}{*}{\textbf{Enzyme}} & \multicolumn{5}{c}{\textbf{MS dataset}} \\
& Organism & Substrate & pH & Temperature & Prediction result \\
\hline
Wildtype
& Cannabis sativa & GPP & 7 & 30 & - \\
M1
& Cannabis sativa & GPP & 7 & 30 & Correct \\
M2
& Cannabis sativa & GPP & 7 & 30 & Correct \\
M3
& Cannabis sativa & GPP & 7 & 30 & Correct \\
M4
& Cannabis sativa & GPP & 7 & 30 & Correct \\
M5
& Cannabis sativa & GPP & 7 & 30 & Incorrect \\
M6
& Cannabis sativa & GPP & 7 & 30 & Correct \\
M7
& Cannabis sativa & GPP & 7 & 30 & Incorrect \\
M8
& Cannabis sativa & GPP & 7 & 30 & Correct \\
\hline
\end{tabular}
\caption{
The zero-shot inference results of MS dataset, where Wildtype is used as a reference.
For each variant, if our model's predicted ranking of its catalytic efficiency relative to Wildtype matches the actual ranking, it is considered a correct prediction; otherwise, it is considered incorrect.
}
\label{table:case-2}
\end{table}

As shown in Table \ref{table:case-1}, the TAL dataset contains the wildtype tyrosine ammonia lyase and its nine variants. 
Using the wildtype as a reference, for each variant, if our model's predicted ranking of its catalytic efficiency ($k_{cat}/K_{m}$) relative to the wildtype matches the actual ranking, then the prediction is considered correct; otherwise, it is considered incorrect.
It is worth noting that the maximum sequence identities of all sequences in the TAL dataset with the training sets of our $k_{cat}$ and $K_{m}$ predictive model is no greater than 59\%.
The prediction results show that our model achieved a success rate of over 77\%, identifying most variants with superior enzyme activity.
For the MS dataset, there is one wildtype myrcene syntheses and its eight variants, with all sequences having the maximum sequence identities with the training sets of less than 51\%.
Our model correctly predicted six of the eight variants, achieving a prediction success rate of 75\%.
Overall, our model exhibits excellent generalization ability for practical enzyme engineering tasks with dissimilar enzymes, thus having broad applications in enzyme discovery and evolution.

\section{Discussion}

In this study, we present $\text{O}^2$DENet, a lightweight and plug-and-play module that improves out-of-distribution generalization in enzyme--substrate interaction prediction by integrating pseudodata augmentation with invariant representation learning.
Without modifying backbone architectures, $\text{O}^2$DENet consistently enhances $k_{cat}$ and $K_{m}$ prediction across multiple representative ESI frameworks and stringent sequence-identity-based OOD splits, indicating that learning perturbation-invariant biochemical features is critical for robust enzyme kinetics modeling.
Biologically and chemically meaningful perturbations, including enzyme sequence masking, substrate SMILES enumeration, and molecular graph masking, expand the effective training distribution, while feature-level consistency regularization suppresses spurious correlations and promotes task-relevant invariance, leading to improved AU-GOOD values and slower performance degradation under increasing distributional shift.
Ablation studies show that moderate masking ratios (approximately 10\%) optimally balance diversity and information preservation, whereas excessive masking appears to disrupt key biochemical signals.
Moreover, zero-shot evaluations on directed evolution datasets demonstrate that $\text{O}^2$DENet generalizes beyond benchmark settings, successfully identifying variants with improved catalytic efficiency despite low sequence similarity to training data. Overall, $\text{O}^2$DENet provides an efficient and broadly applicable strategy for enhancing the reliability and deployability of data-driven enzyme kinetics prediction models, with direct implications for enzyme discovery and engineering.

\section{Methods}

\subsection{Datasets}

Two benchmark datasets from CatPred-DB \cite{CatPred}, namely CatPred-$k_{cat}$ and CatPred-$K_{m}$ datasets, were adopted in this work.
Each record in the above datasets represents an enzyme–substrate pair, consisting of an enzyme sequence with corresponding structure, a substrate SMILES~\cite{weininger1988smiles}, and the corresponding kinetic parameter ($k_{cat}$ or $K_{m}$). 
All enzymes were standardized according to UniProt annotations, and substrate SMILES were canonicalized using RDKit~\cite{landrum2025rdkit}.
To assess out-of-distribution (OOD) generalization of different models, we adopted the sequence-identity-based dataset splits rather than random splits. 
Specifically, enzyme sequences were clustered using mmseqs2~\cite{steinegger2017mmseqs2} at identity thresholds of 40\%, 60\%, 80\% and 99\% to create different OOD-level splits. In the pseudodata augmentation stage, we generated an augmented dataset with the same order of magnitude as the original training data.
This design choice was made to ensure proper alignment with the task-specific loss, preventing the augmented samples from disproportionately dominating the optimization process.

\subsection{Baseline ESI Frameworks}

To evaluate the universality and effectiveness of the proposed module, we integrated it with four representative enzyme–substrate interaction (ESI) predictive frameworks: UniKP, DLKCat, CatPred, and OmniESI~\cite{UniKP, DLKCat, CatPred, OmniESI}. 
These models cover a range of commonly used architectures, including convolutional neural network, graph neural network, and Transformer-based protein encoders, as well as molecular encoders based on SMILES Transformer~\cite{smilestransformer} and graph neural network~\cite{kroll2023turnover}. 
Such diversity provides a robust basis for assessing the plug-and-play compatibility of our method across different feature extraction paradigms.
For all baselines, our module was incorporated without modifying the backbone network or loss design; only an additional regularization term was applied during training. 
This design allows direct comparison with the original frameworks and highlights the generalization improvement brought by our input augmentation and feature-consistency strategies.

\subsection{Enzyme-substrate-pair Augmentation}

To improve the robustness of ESI predictors against distributional shifts, we designed a pseudo-enzyme-substrate augmentation strategy that applies biologically and chemically meaningful perturbations to both enzymes and substrates. 
Ultimately, we obtained pseudo-enzyme-substrate pairs on the same scale as the original ones.
This strategy aims to increase data diversity while maintaining the biochemical integrity of each enzyme–substrate pair, ensuring that augmented samples are as identical as possible to the original samples in terms of catalytic interactions.

\subsubsection{Enzyme Augmentation} 
For enzyme sequences, we applied a residue-level random masking strategy to introduce controlled perturbations. 
Specifically, given an enzyme sequence $\mathbf{E} = [e_1, e_2, ..., e_n]$, a proportion $p_s$ of amino acids is selected and replaced with a special [MASK] token. 
This strategy encourages the model to rely less on specific individual residues and instead learn more robust, context-aware representations that capture global sequence characteristics.
We evaluated masking ratios of 5\%, 10\%, 15\%, 20\%, 25\%, and 30\% in the ablation study, and found that a ratio of 10\% provides the best balance between effective perturbation and preservation of informative sequence patterns.

\subsubsection{Substrate Augmentation} 
For substrate molecules, we combined two perturbation strategies, namely SMILES enumeration and molecular graph masking, to accommodate different encoders of different baseline ESI frameworks.

\paragraph{SMILES Enumeration} 
Since a molecule can be represented by multiple valid SMILES strings that correspond to the same chemical graph, randomizing atom traversal orders provides diverse yet chemically equivalent textual inputs. 
This augmentation helps the model avoid overfitting to the canonical SMILES format and focus instead on learning molecular topology.

\paragraph{Graph Masking}
Each substrate molecule was represented as a 2D molecular graph $G=(V,E)$, where nodes represent atoms and edges represent bonds. 
A proportion $p_g$ of nodes or edges was randomly masked and replaced by a learnable embedding vector, while ensuring that atoms within functional groups, ring systems, and reactive centers were excluded from masking. 
This procedure introduces controlled structural perturbations without altering the molecule’s core reactivity or functional identity. 
We evaluated graph masking ratios of 5\%, 10\%, 15\%, 20\%, 25\%, and 30\%~\cite{graphmask}, and observed the highest performance at 10\%, beyond which excessive masking degraded prediction performance.

\subsection{Training Objective}

The overall training objective integrated the task-specific loss $\mathcal{L}_{\text{base}}$ with a feature consistency regularization term $\mathcal{L}_{\text{cons}}$, which was inspired by invariance-based out-of-distribution generalization techniques.
The goal is to optimize for both accurate kinetic predictions and robust, perturbation-invariant feature representations that generalize well across distribution shifts.

\paragraph{Feature Consistency Regularization}

Invariant representation learning aims to extract task-relevant features that remain stable under input perturbations or environmental variations, thereby improving model robustness and OOD generalization.
For each enzyme–substrate pair $(E, S)$ and its augmented counterpart $(E', S')$, the model extracted feature representations $\mathbf{f}_{ES}$ and $\mathbf{f}'_{ES}$, respectively.
We applied a feature consistency loss to enforce invariant latent representations between each sample (the raw ESI embedding here) and its pseudo-data variant (the augmented ESI embedding here).

The consistency regularization term is formulated as:
\begin{equation}
\mathcal{L}_{\text{cons}} = \|\mathbf{f}_{ES} - \mathbf{f}'_{ES}\|_2^2,
\end{equation}
where $\mathbf{f}_{ES}$ and $\mathbf{f}'_{ES}$ are the representations of the raw and augmented enzyme–substrate pairs.
This loss encourages the model to learn features that are invariant to augmentations, such as sequence masking and graph masking.

\paragraph{Task-Specific Loss}

For the primary downstream task of enzyme–substrate interaction prediction, we used a regression loss function $\mathcal{L}_{\text{base}}$, typically mean squared error (MSE), to predict kinetic parameters such as $k_{cat}$ or $K_{m}$:
\begin{equation}
\mathcal{L}_{\text{base}} = \frac{1}{N} \sum_{i=1}^{N} (y_i - \hat{y}_i)^2,
\end{equation}
where $y_i$ and $\hat{y}_i$ are the experimental (ground-truth) and predicted kinetic values, respectively, and $N$ is the number of samples.

\paragraph{Total Training Objective}

The total training objective combines both the contrastive regularization term and the primary downstream task loss:
\begin{equation}
\mathcal{L}_{\text{total}} = \mathcal{L}_{\text{base}} + \lambda \, \mathcal{L}_{\text{cons}},
\end{equation}
where $\lambda$ is a hyperparameter that controls the trade-off between the primary loss $\mathcal{L}_{\text{base}}$ and the consistency regularization $\mathcal{L}_{\text{cons}}$.

The above total objective encourages the model to make accurate predictions on enzyme–substrate interactions while ensuring that the learned representations are stable and invariant under different augmentations.
The feature consistency loss $\mathcal{L}_{\text{cons}}$ effectively acts as a self-supervised regularization that forces the model to focus on the most informative and generalizable features of the enzyme–substrate pairs. 
By incorporating this regularization, the model becomes less sensitive to noise introduced by data augmentations, thereby improving its robustness to OOD data and enhancing its generalization ability.
Hyperparameter $\lambda$ was tuned over a range of $[0.5, 1.0]$ based on the validation performance to find the optimal balance between predictive accuracy and feature robustness.

\subsection{Evaluation Metrics}

We employed three complementary metrics to assess model performance: the coefficient of determination ($R^2$), mean absolute error (MAE), and the AU-GOOD (Area Under the GOOD Curve) metric~\cite{AUGOOD}. 
While $R^2$ and MAE quantify overall prediction accuracy, AU-GOOD evaluates the model’s expected reliability under distributional shifts.

\paragraph{Coefficient of Determination}
The $R^2$ metric measures how well the predicted values $\hat{y}_i$ approximate the experimental values $y_i$:
\begin{equation}
R^2 = 1 - \frac{\sum_{i=1}^{N} (y_i - \hat{y}_i)^2}{\sum_{i=1}^{N} (y_i - \bar{y})^2},
\end{equation}
where $\bar{y}$ is the mean of the observed data and $N$ is the number of samples.

\paragraph{Mean Absolute Error}
MAE reflects the average absolute deviation between predicted and experimental values:
\begin{equation}
\mathrm{MAE} = \frac{1}{N}\sum_{i=1}^{N} |y_i - \hat{y}_i|.
\end{equation}

\paragraph{Area Under the GOOD Curve (AU-GOOD)}

The AU-GOOD metric estimates the expected empirical risk of a model conditioned on the target deployment distributions. 
It integrates model performance across similarity thresholds between training and test data, weighted by how likely each similarity level appears in the target distribution~\cite{AUGOOD}.
Formally:
\begin{equation}
\mathrm{AU\text{-}GOOD} = 
\int_{\lambda_0}^{\lambda_n} 
\hat{R}_{\lambda_s}(f_\theta)\,P(\lambda_s|P^*)\,d\lambda_s
\approx
\sum_i 
\hat{R}_{\lambda_i}(f_\theta)\,
P(\lambda_i|P^*)\,\Delta \lambda_s,
\end{equation}
where $\hat{R}_{\lambda_s}(f_\theta)$ is the empirical risk at similarity threshold $\lambda_s$, 
and $P(\lambda_s|P^*)$ denotes the histogram of maximal similarities between target and training samples.

Intuitively, AU-GOOD summarizes how model performance degrades as the similarity between training and deployment data decreases. 
A higher AU-GOOD value indicates better generalization to unseen or out-of-distribution regions, capturing not only accuracy but also robustness and stability across similarity shifts.

\subsection{Implementation}

All models were deployed on an Ubuntu 20.04 platform equipped with an Intel(R) Xeon(R) Gold 6430 CPU and an NVIDIA RTX 4090 GPU.
The system was configured with up to 120 GB of RAM, and all models were developed using PyTorch (Python 3.7.13).

\section*{Data and code availability}
The CatPred-$k_{cat}$ dataset, CatPred-$K_{m}$ dataset, the directed evolution datasets for tyrosine ammonia lyase (TAL) and myrcene synthase (MS) are freely available from previous studies \cite{CatPred,mpek}
Relevant data, code, and models of this work are available via GitHub at \url{https://github.com/blackjack534/AuESINet}.

\bibliography{sn-bibliography}

\section*{Acknowledgments}

This work was supported by Guangdong S\&T Programme (Grant No. 2024B0101010003).

\section*{Competing interests}

The authors declare no competing interests.

\newpage
\section*{Supplementary information}

\setcounter{section}{1}
\renewcommand{\thesubsection}{S\arabic{section}}
\subsection{Ablation study for masking ratios}

\newpage

\newpage

\newpage
\setcounter{section}{2}
\renewcommand{\thesubsection}{S\arabic{section}}
\subsection{Hyperparameter sensitivity analysis for weighting coefficient $\lambda$}

In this section, we conducted a hyperparameter sensitivity analysis with respect to the weighting coefficient $\lambda$ introduced in the loss function.
All experiments were performed on the CatPred-$K_m$ dataset, where five values of $\lambda$ spanning multiple orders of magnitude were evaluated.
Model performance was assessed using both $R^2$ and MAE on different OOD test splits.
As reported in Table~\ref{tabs:lambda_sensitivity_placeholder}, the model achieved the best overall performance when $\lambda = 0.5$.

\begin{table}[h!]
\centering
\renewcommand{\thetable}{S1}
\begin{tabular}{c|cccc|cccc}
\hline
\multirow{2}{*}{$\lambda$}
& \multicolumn{4}{c|}{$R^2 \uparrow$}
& \multicolumn{4}{c}{$\mathrm{MAE} \downarrow$} \\
& OOD-99\% & OOD-80\% & OOD-60\% & OOD-40\%
& OOD-99\% & OOD-80\% & OOD-60\% & OOD-40\% \\
\hline
0.005 & 0.558 & 0.567 & 0.536 & 0.501 & 0.628 & 0.642 & 0.673 & 0.720 \\
0.05  & 0.541 & 0.579 & 0.547 & 0.515 & 0.612 & 0.625 & 0.662 & 0.718 \\
\textbf{0.5}  & \textbf{0.576} & \textbf{0.584} & \textbf{0.551} & \textbf{0.519}
              & \textbf{0.606} & \textbf{0.618} & \textbf{0.646} & \textbf{0.692} \\
5     & 0.570 & 0.581 & 0.546 & 0.513 & 0.611 & 0.622 & 0.651 & 0.700 \\
50    & 0.561 & 0.572 & 0.538 & 0.504 & 0.623 & 0.636 & 0.668 & 0.716 \\
\hline
\end{tabular}
\caption{Hyperparameter sensitivity analysis of the weighting coefficient $\lambda$ for the $K_m$ prediction task.
Performance was evaluated using $R^2$ and MAE across different OOD test splits.
}
\label{tabs:lambda_sensitivity_placeholder}
\end{table}

\end{document}